\documentclass[11pt,a4paper]{article}
\usepackage[hyperref]{acl2019}
\usepackage{times}
\usepackage{latexsym}
\usepackage{graphicx}
\usepackage{algorithmic}
\usepackage{url}
\usepackage{multirow}

\aclfinalcopy % Uncomment this line for the final submission

\title{Adversarial Attacks and Defense on Texts: A Survey}

\author{Aminul Huq \& Mst. Tasnim Pervin\\
  ID: 2019280161 \& 2019280162\\
  Dept. of Computer Science \& Technology \\
  Tsinghua University\\
    \texttt{\{huqa10,pervinmt10\}@mails.tsinghua.edu.cn }
   \\}

\date{}

\begin{document}
\maketitle
\begin{abstract}
  
Deep learning models have been used widely for various purposes in recent years in object recognition, self-driving cars, face recognition, speech recognition, sentiment analysis, and many others. However, in recent years it has been shown that these models possess weakness to noises which force the model to misclassify. This issue has been studied profoundly in the image and audio domain. Very little has been studied on this issue concerning textual data. Even less survey on this topic has been performed to understand different types of attacks and defense techniques. In this manuscript, we accumulated and analyzed different attacking techniques and various defense models to provide a more comprehensive idea. Later we point out some of the interesting findings of all papers and challenges that need to be overcome to move forward in this field.
  
\end{abstract}

\section{Introduction}

From the beginning of the past decade, the study and application of Deep Neural Network (DNN) models have sky-rocketed in every research field. It is currently being used for computer vision \cite{buch2011review,borji2014salient}, speech recognition\cite{deng2013recent, huang2015analysis}, medical image analysis\cite{litjens2017survey,shen2017deep}, natural language processing\cite{zhang2018deep,otter2020survey}, and many more. DNN’s are capable of solving large scale complex problems with relative ease which, tends to be difficult for regular statistical machine learning models. That is why in many real-world applications, DNN’s are used profoundly and explicitly. In a study by C. Szegedy\shortcite{szegedy2013intriguing}, showed that DNN models are not that robust in the image domain. They are in fact, quite easy to fool and can be tampered in such a way to obey the will of an adversary. This study caused an uproar in the researcher community and researchers started to explore this issue in other research areas as well. Different researchers worked tirelessly and showed that DNN models were vulnerable in object recognition systems \cite{goodfellow2014explaining}, audio recognition\cite{carlini2018audio}, malware detection \cite{grosse2017adversarial}, and sentiment analysis systems \cite{ebrahimi2017hotflip} as well. An example of the adversarial attacks is shown in figure 1.

The number of studies of adversarial attacks and defenses in the image domain outnumbers the number of studies performed in textual data \cite{wang2019survey}. In Natural Language Processing (NLP), for various applications like sentiment analysis, machine translations, question-answering, and in many others, different attacks and defense have been employed. In the field of NLP, Papernot \shortcite{papernot2016crafting} paved the way by showing that adversarial attacks can be implemented for textual data as well. After that, various research has been performed to explore adversarial attacks and defense in the textual domain.

Adversarial attacks are a security concern for all real-world applications that are currently running on DNNs. It is the same scenario for NLP as well. There are many real-world programs launched, which is based on DNNs like sentiment analysis \cite{pang2008opinion}, text question-answering \cite{gupta2012survey}, machine translation \cite{wu2016google}, and many others. Users in the physical world use these applications in their lives to get suggestions about a product, movies, or restaurant or to translate texts. An Adversary could easily use attack techniques for ill-will and provide wrong recommendations and falsify texts. Since these attacks are not observed by the models due to the lack of robustness these programs would lose values. Thus, the study of adversarial attacks and defense with respect to text is of utmost importance. 

\begin{figure}[ht]
\centering
  \includegraphics[width=7cm,height=4.5cm]{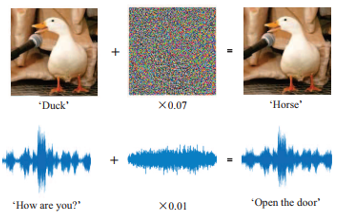}
\caption{Adversarial attacks on image and audio data \cite{gong2018protecting}}
\label{fig:1}       % Give a unique label
\end{figure}

In this review, our major contributions can be summarized as follows.

\begin{itemize}
    \item We provide a systematic analysis and study of different adversarial attacks and defense techniques which are shown in different research projects related to classification, machine translations, question-answering, and many others.
    \item We present here adversarial attack and defense techniques by considering different attack levels.
    \item After going through all the research works we have tried to answer which attack and defense technique has the advantage over other techniques.
    \item Finally, we present some exciting findings after going through all the research works and point out challenges that need to be overcome in the future.
\end{itemize}

Our manuscript is organized as follows. Related research works are mentioned in section $2$. In section $3$ we start by providing preliminary information about adversarial machine learning concerning both image and textual data. Here we also provide a classification of different attack and defense approaches as well. Following this, in section $4$ we discuss various attack techniques based on the taxonomy presented in section $3$. To defend models from these attack techniques we discuss different defense techniques in section $5$. We provide an in-depth discussion of our findings in these topics and some challenges in section $6$. We conclude our manuscript by providing a conclusion in section $7$.

\section{Related Works}

For textual data this topic has not been explored that much but for the image domain, it has been explored much more. Since this topic hasn’t been explored that much small number of publications about it have been found and a smaller number of review works have been done. We were able to go through three review papers related to this topic \cite{belinkov2019analysis,xu2019adversarial,zhang2020adversarial}. In the manuscript of Belinkov\shortcite{belinkov2019analysis}, they mainly studied different analysis in NLP, visualization and what type of information neural networks capture. They introduced adversarial examples to explain that traditional DNN models are weak. Xu \shortcite{xu2019adversarial} explained adversarial examples in all domains i.e. image, audio, texts etc. It was not specialized for text but mostly related to image domain. Zhang \shortcite{zhang2020adversarial} in their manuscript described different publications related to adversarial examples. Unfortunately it was focused largely on attack strategies and shed little light on defense technique. They mainly discussed data augmentation, adversarial training and one distillation technique proposed by \cite{papernot2016crafting}. Table 1 provides a comparative analysis between these survey papers and ours.

\begin{table}[t]
\begin{tabular}{|c|c|c|c|}
\hline
         & \begin{tabular}[c]{@{}c@{}}Text \\ Domain\end{tabular} & Attacks                & Defense            \\ \hline
Belinkov\shortcite{belinkov2019analysis} & \multirow{2}{*}{Partly}                                & \multirow{4}{*}{*****} & \multirow{2}{*}{*} \\ \cline{1-1}
Xu\shortcite{xu2019adversarial}       &                                                        &                        &                    \\ \cline{1-2} \cline{4-4} 
Zhang\shortcite{zhang2020adversarial}    & \multirow{2}{*}{Fully}                                 &                        & **                 \\ \cline{1-1} \cline{4-4} 
Ours     &                                                        &                        & ****               \\ \hline
\end{tabular}
\caption{Comparison with recent surveys. (No. of stars represent how much that topic is discussed)}
\label{tab:my-table1}
\end{table}

\section{Adversarial Machine Learning}

Modern machine learning and deep learning has achieved a whole new height because of its high
computational power and fail-proof architecture. However, recent advances in adversarial training have broken this illusion. A compelling model can misbehave by a simple attack by adversarial examples. An adversarial example is a specimen of input data that has been slightly transformed in such a way that can fool a machine learning classifier resulting in misclassification. The main idea behind this attack is to inject some noise to the input to be classified that is unnoticeable to the human eye so that the resulting prediction is changed from actual class to another class. Thus we can understand the threat of this kind of attack to classification models.

\subsection{Definition}

For a given input data and its labels $(x,y)$ and a classifier $F$ which is capable of mapping inputs $x$ to its designated labels $y$ in the general case we can define them as $F(x) = y$. 
However, for adversarial attack techniques apart from input data a small perturbation $\delta$ is also added to the classifier $F$. Note that, this perturbation is imperceptible to human eyes and it is limited by a threshold $||\delta||<\epsilon$. In this case, the classifier is unable to map it to the original labels. Hence, $F(x+\delta) \neq y$.
The concept of imperceptibility is discussed in length in section $5.2$.

A robust DNN model should be able to look beyond this added perturbation and be able to classify input data properly. i.e. $F(x+\delta) = y$.

\subsection{Existence of Adversarial Noises}

Since adversarial examples have been uncovered, a growing and difficult question have been looming over the research community. Why adversarial examples exist in real-life examples. Several hypotheses have been presented in an attempt to answer this question. However, none of them have achieved a unanimous agreement of the overall researcher community. The very first explanation comes from C. Szegedy’s paper \shortcite{szegedy2013intriguing}. In it, he says that adversarial examples exist because there are too much non-linearity and the network is not regularized properly. Opposing this hypothesis I. Goodfellow says that it exists because of too much linearity in machine learning and deep learning models \shortcite{goodfellow2014explaining}. Various activation functions that we use today like ReLU and Sigmoid are straight lines in the middle parts. He argues that since we want to protect our gradients from vanishing or exploding we tend to keep our activation functions straight. Hence, if a small noise is added to the input because of the linearity it perpetuates in the same direction and accumulates at the end of the classifier and produces miss-classification. Another hypothesis that is present today is called tilted boundary \cite{tanay2016boundary}. Since the classifier is never able to fit the data exactly there is some scope for the adversarial examples to exist near the boundaries of the classifier. A recent paper argues that adversarial examples are not bugs but they are features and that is how deep neural networks visualize everything \cite{ilyas2019adversarial}. They classified the features into two categories called robust and non-robust features and showed that by adding small noises non-robust features can make the classifier to provide a wrong prediction.

\subsection{Classification of Adversarial Examples}

Here, we provide a basic taxonomy of adversarial attacks and adversarial defense techniques based on different metrics. 
For adversarial attack techniques, we can classify different attacks based on how much the adversary has knowledge about the model. We can divide it into two types.

\begin{itemize}
    \item White-Box Attacks: In order to execute these types of attacks the adversary needs to have full access to the classifier model. Using the model parameters, architectures, inputs, and outputs the adversary launch the attack. These types of attacks are the most effective and harmful since it has access to the whole model.
    \item Black-Box Attacks: These attacks represent real-life scenarios where the adversary has no knowledge about the model architecture. They only know about the input and output of the model. In order to obtain further information, they use queries.
\end{itemize}

We can classify attacks based on the goal of the adversary as well. We can classify it into two types.

\begin{itemize}
    \item Non-Targeted Attack: Adversary in this scenario does not care about the labels that the model produces. They are only interested in reducing the accuracy of the model. i.e. $F(x+\delta) \neq y$.
    \item Targeted Attack: In targeted attacks, the adversary forces the model to produce a specific output label for given images. i. e. $F(x+\delta) = y*$.
\end{itemize}

These are the general classifications of different attacks. However, for NLP tasks, we can classify attacks differently. Since the data in text-domain is different from the data in the image or audio domain attack strategy and attack types are somewhat different. Based on which components are modified in the text we can classify attack techniques into four different types. They are called character-level attacks, word-level attacks, sentence-level attacks, and multi-level attacks. In these adversarial attacks text data are generally inserted, removed, swapped/replaced, or flipped. Though not all of these options are explored in different levels of attacks. 

\begin{itemize}
    \item Character-Level Attack: Individual characters in this attack are either modified with new characters, special characters, and numbers. These are either added to the texted, swapped with a neighbor, removed from the word, or flipped.
    \item Word-Level Attack: In this attacks words from the texts are changed with their synonyms, antonyms, or changed to appear as a typing mistake or removed completely.
    \item Sentence-Level Attack: Generally new sentences are inserted as adversarial examples in these types of attacks. No other approach has been explored yet. 
    \item Multi-Level Attack: Attacks which can be used in a combination of character, word, and sentence level are called multi-level attack.
\end{itemize}

\begin{figure}[ht]
\centering
  \includegraphics[width=7cm]{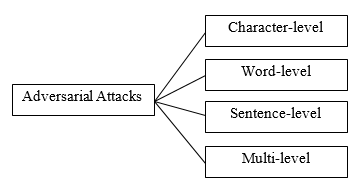}
\caption{Adversarial attacks classification}
\label{fig:2}       % Give a unique label
\end{figure}

Adversarial defense techniques have been studied in mainly three directions \cite{akhtar2018threat}. They are

\begin{itemize}
    \item Modified Training/Input: In these cases, the model is trained differently to learn more robust features and become aware of adversarial attacks. During testing, inputs are also modified to make sure no adversarial perturbation is added to it.
    \item Modifying Networks: By adding more layers or sub-networks and changing loss or activation functions defense is sought in this scenario.
    \item Network Add-on: Using external networks as additional sections for classifying unseen data. 
\end{itemize}

\section{Adversarial Attacks}
Most of the literature is about attack techniques that are where we start our discussion. In this section, we will be analyzing different attack techniques published in recent years in detail. In order to provide a clear understanding, we are diving our explanations based on the taxonomy for the NLP that we mentioned in section 2.

\subsection{Character-Level Attack}

\begin{figure*}[ht]
  \includegraphics[width=16cm,height=2.75cm]{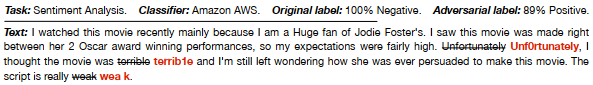}
\caption{Example of character-level attack(TEXTBUGGER) \cite{li2018textbugger}}
\label{fig:3}       % Give a unique label
\end{figure*}

\begin{table}
\begin{tabular}{|c|c|c|}
\hline
Reference & Attack-type                                                          & Application                                                    \\ \hline
Ebrahimi\shortcite{ebrahimi2018adversarial} & \begin{tabular}[c]{@{}c@{}}White-box \\ and\\ Black-box\end{tabular} & \begin{tabular}[c]{@{}c@{}}Machine \\ Translation\end{tabular} \\ \hline
Belinkov\shortcite{belinkov2017synthetic}  & Black-box                                                            & \begin{tabular}[c]{@{}c@{}}Machine \\ Translation\end{tabular} \\ \hline
Gao\shortcite{gao2018black}      & Black-box                                                            & Classification                                                 \\ \hline
Li\shortcite{li2018textbugger}        & \begin{tabular}[c]{@{}c@{}}White-box \\ and\\ Black-box\end{tabular} & Classification                                                 \\ \hline
Gil\shortcite{gil2019white}       & Black-box                                                            & Classification                                                 \\ \hline
Hosseini\shortcite{hosseini2017deceiving}  & Black-box                                                            & Classification                                                 \\ \hline
\end{tabular}
\caption{Character-level attack type with applications.}
\label{tab:my-table2}
\end{table}

As mentioned before character level attacks includes attack schemes which try to insert, modify, swap, or remove a character, number, or special character. Ebrahimi \shortcite{ebrahimi2018adversarial} in his paper worked with generating adversarial examples for character-level neural machine translation. They provided white and black box attack techniques and showed that white-box attacks were more damaging than black-box attacks. They proposed a controlled adversary that tried to mute a particular word for translation and targeted adversary which aimed to push a word into it. They used gradient-based optimization and in order to edit the text, they performed four operations insert, swap two characters, replace one character with another and delete a character. For black-box attack, they just randomly picked a character and made necessary changes. 

Belinkov \shortcite{belinkov2017synthetic} worked with character-based neural machine translation as well. In their paper, they didn’t use or assume any gradients. They relied on natural and synthetic noises for generating adversarial noises. For natural noises, they collected different errors and mistakes from various datasets and replaced correct words with wrong ones. In order to generate synthetic noises they relied on swapping characters, randomized characters of a word except the first and last one, randomized all the characters, and replaced one character with a character from its neighbor in the keyboard.

Another black-box attack was proposed by Gao \shortcite{gao2018black}. They worked with Enron spam emails and IMBD dataset for classification tasks. Since in the black-box settings, an adversary does not have access to gradients they proposed a two-step process to determine which words are the most significant ones. Temporal score and temporal tail scores are to be calculated to determine the most significant word. This approach was coined as DEEPWORDBUG by the authors. To calculate the temporal score, they checked how much effect each word had on the classification result. The Temporal tail score is the complement of temporal scores.  For temporal tail score, they compared results for two trailing parts of sentences, one had a particular word and another did not have it.

TEXTBUGGER is both a white-box and black-box attack framework that was proposed by Li \shortcite{li2018textbugger}. For generating bugs or adversarial examples they focused on five kinds of edits: insertion, deletion, swapping, substitution with a visually similar word, substitution with semantically similar meaning. For white-box attacks, they proposed a two-step approach. The first step is to determine which words are most significant with the help of determining the Jacobian matrix. Then generate all five bugs and choose the one which is the most optimal for reducing accuracy. In order to generate black-box attacks in this framework, they propose a three-step approach. Since there is no access to the gradient thus they propose to determine first which sentence is the most important one. Then determine which word is the most significant and finally generate five bugs for it and choose which one is the most optimal.

Gil \shortcite{gil2019white} was able to transform a white-box attack technique to a black-box attack technique. They generated adversarial examples from a white-box attack technique and then trained a neural network model to imitate the overall procedure. They transferred the adversarial examples generation by the HotFlip approach to a neural network. They coined these distilled models as DISTFLIP. Their approach had the advantages of not being dependent on the optimization process which made their adversarial example generation faster.

Perspective is an API that is built by Google and Jigsaw to detect toxicity in comments. Hosseini \shortcite{hosseini2017deceiving} showed that it can be deceived by modifying inputs. They didn’t have any calculated approach, mainly modified toxic words by adding a dot (.) or space between two characters or swapping two characters, and thus they showed the API got lower toxicity score than before.

\subsection{Word Level Attack}

\begin{figure*}[ht]
  \includegraphics[width=16cm,height=3.5cm]{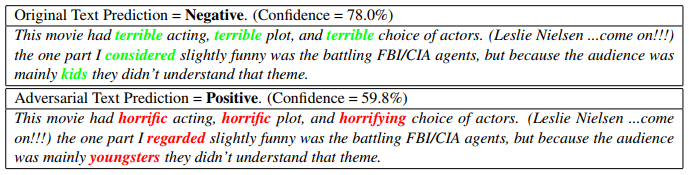}
\caption{Example of word-level attack \cite{alzantot2018generating}}
\label{fig:4}       % Give a unique label
\end{figure*}

\begin{table}
\begin{tabular}{|c|c|c|}
\hline
Reference & Attack-type & Application    \\ \hline
Papernot\shortcite{papernot2016crafting}  & White-box   & Classification \\ \hline
Samanta\shortcite{samanta2017towards}   & White-box   & Classification \\ \hline
Liang\shortcite{liang2017deep} & \begin{tabular}[c]{@{}c@{}}White-box \\ and\\ Black-box\end{tabular} & Classification \\ \hline
Alzantot\shortcite{alzantot2018generating}  & Black-box   & Classification \\ \hline
Kulesov\shortcite{kuleshovadversarial}   & White-box   & Classification \\ \hline
Zang\shortcite{zang2019textual}      & Black-box   & Classification \\ \hline
\end{tabular}
\caption{Word-level attack type with applications.}
\label{tab:my-table3}
\end{table}

Papernot \shortcite{papernot2016crafting} was the first one to generate adversarial examples for texts. They used a computational graph unfolding technique to calculate the forward derivative and with its help the Jacobian. It helps to generate adversarial examples using the FGSM technique. The words they choose to replace with are chosen randomly so the sentence doesn’t keep original meaning or grammatical correctness.

To change a particular text classification label with the minimum number of alterations Samanta \shortcite{samanta2017towards} proposed a model. In their model, they either inserted a new word or deleted one or replaced one. They at first determined which words are highly contributing to the classifier. They determined a word is highly contributing if removing it changes the class probability to a large extend. To replace the words they created a candidate pool based on synonyms, typos that produce meaningful words and genre-specific words. They changed a particular word based on the following conditions.

\begin{algorithmic}
\IF {Word is an adverb and highly contributing} 
    \STATE remove it
\ELSE
    \STATE Choose a word from the candidate pool
        \IF {the word is an adjective and candidate word is adverb\\}
            \STATE Insert
        \ELSE
             \STATE replace particular word with candidate word
        \ENDIF
\ENDIF 
\end{algorithmic}

Liang \shortcite{liang2017deep} proposed a white-box and black-box attack strategy based on insertion, deletion, and modification. To generate adversarial examples they used natural language watermarking technique \cite{atallah2001natural}. To perform a white-box attack they provided a concept of Hot Training Phrase (HTP) and Hot Sample Phrase (HSP). These are obtained with the help of back-propagation to get all the cost gradients of each character. HTP helps to determine what needs to be inserted while HSP helps to determine where to insert, delete, and modify. For black-box attacks, they borrowed the idea of fuzzing technique \cite{sutton2007fuzzing} for implementing a test to get HTPs and HSPs.

To preserve syntactical and semantic meaning Kuleshov  \shortcite{kuleshovadversarial} used thought vectors. They took the inspiration from \cite{bengio2003neural,mikolov2013efficient} which mapped sentences to vectors. Those who had similar meanings were placed together. To ensure semantic meaning they introduce syntactic constraint. Their approach was iterative and in each iteration, they replaced only one word with their nearest neighbor which changed the objective function the most.  

Genetic algorithm based black-box attack techniques are proposed by Alzanot \shortcite{alzantot2018generating}. They tried to generate adversarial examples that were semantically and syntactically similar. For a particular sentence, they randomly select a word and replace it with a suitable replacement word that fits the context of the sentence. For this, they calculate the first few nearest neighbor words according to the GloVe embedding space. Next using Google’s 1 billion words language model they try to remove any words which do not match the context. After that, they select a particular word which maximizes the predication. This word is then inserted into the sentence. 

An improvement of the genetic algorithm based attack was proposed by Wang \shortcite{wang2019natural}. They modified it by allowing a single word of a particular sentence to be changed multiple times. To ensure that the word is indeed a synonym of the original word it needs to be fixed.

A sememe based word substitution method using particle swarm optimization technique was proposed by \cite{zang2019textual}. A sememe is the minimum semantic unit in human language. They argued that word embedding and language model based substitution methods can find many replacements but they are not always semantically correct or related to the context. They compared their work with \cite{alzantot2018generating} attack technique and showed their approach was better. 

\subsection{Sentence Level Attack}

\begin{table}
\begin{tabular}{|c|c|c|}
\hline
Reference & Attack-type & Application                                                             \\ \hline
Jia\shortcite{jia2017adversarial} & \begin{tabular}[c]{@{}c@{}}White-box\\ and\\ Black-box\end{tabular} & \begin{tabular}[c]{@{}c@{}}Question\\ Answering\end{tabular} \\ \hline
Wang\shortcite{wang2018robust}      & White-box   & Classification                                                          \\ \hline
Zhao\shortcite{zhao2017generating}      & Black-box   & \begin{tabular}[c]{@{}c@{}}Natural \\ Language\\ Inference\end{tabular} \\ \hline
Cheng\shortcite{cheng2019robust}     & White-box   & \begin{tabular}[c]{@{}c@{}}Machine\\ Translation\end{tabular}           \\ \hline
Micheal\shortcite{michel2019evaluation}   & White-box   & \begin{tabular}[c]{@{}c@{}}Machine\\ Translation\end{tabular}           \\ \hline
\end{tabular}
\caption{Sentence-level attack types with applications.}
\label{tab:my-table4}
\end{table}

In the domain of question answering Robin Jia \shortcite{jia2017adversarial} introduced two attack techniques called ADDSENT and ADDANY. They also introduced two variants of this ADDONESENT and ADDCOMMON randomly. Here, ADDONESENT is a model-independent attack i.e. black-box attack. Using these attacks they generated an adversarial example that does not contradict the original answer and inserts it at the end of the paragraph. To show the effectiveness of ADDSENT and ADDANY they used it on 16 different classifiers and showed that all of them got reduced F1 score. Figure 5. provides a visual representation of how ADDSENT and ADDANY attack is used for generating texts.

\begin{figure*}[ht]
  \includegraphics[width=16cm,height=8cm]{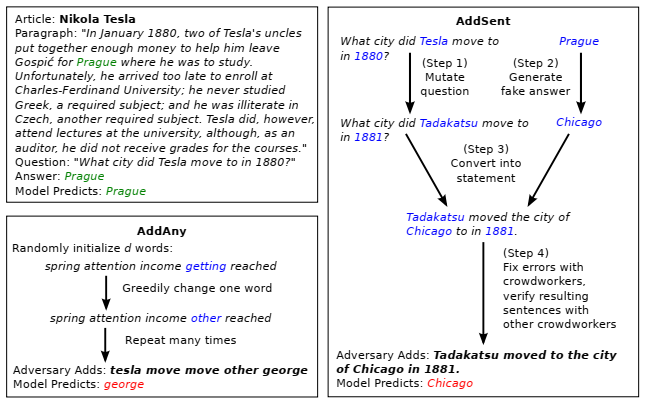}
\caption{ADDANY and ADDSENT Attack Generation \cite{jia2017adversarial}}
\label{fig:5}       % Give a unique label
\end{figure*}

There is another group of researchers named Yicheng Wang \shortcite{wang2018robust} who have worked on the modification of the ADDSENT model. They proposed two modifications of the ADDSENT model and named their model as ADDSENTDIVERSE. ADDSENT model creates fake answers that are semantically irrelevant but follows similar syntax as a question. In ADDSENTDIVERSE, they targeted to generate adversarial examples with a higher variance where distractors will have randomized placements so that the set of fake answers will be expanded. Moreover, to address the antonymstyle semantic perturbations that are used in ADDSENT, they added semantic relationship features enabling the model to identify the semantic relationship among contexts of questions with the help of WordNet. The paper shows that the ADDSENTDIVERSE model beats ADDSENT trained model by an average improvement of 24.22\% in F1 score across three different classifiers indicating an increase in robustness.

Zhao \shortcite{zhao2017generating} Proposed a new framework utilizing Generative Adversarial Networks (GAN) on Stanford Natural Language Interface (SNLI) dataset to generate grammatically legible and natural adversarial examples that are valid and semantically close to the input and can detect local behavior of input by searching in semantic space of continuous data representation. They have implemented these adversaries in different applications such as image classification, machine translation, and textual entailment to evaluate the performance of their proposed approach on black-box classifiers such as  ARAE ( Adversarially Regularized Autoencoder), LSTM and TreeLSTM. By their work, they have proved that their model is successful to generate adversaries that can pass common-sense reasoning by logical inference and detect the vulnerability of the Google Translate model during machine translation.

Cheng \shortcite{cheng2019robust} worked with neural machine translation and proposed a gradient-based white-box attack technique called AdvGen. Guided by the training loss they used a greedy choice based approach to find the best solution. They also used the language model into it as well because it is computationally easy for solving an intractable solution and it also retains somewhat semantic meaning. Their research paper is based on using adversarial examples for both attack generation and using these adversarial examples to improve the robustness of the model.

Michael\shortcite{michel2019evaluation} worked with neural machine translation as well and in their manuscript, they proposed a natural criterion for untargeted attacks. It is “adversarial examples should be meaning preserving on the source side but meaning destroying on the target side”. From it, we can see that they are focusing on the point about preserving the meaning of the sentences while pushing adversarial examples into it. They propose a white-box attack using the gradients of the model which replaces one word from the sentences to maximize the loss. To preserve the meaning of the sentences they used KNN to determine the top 10 words which are similar to a given word. This approach has the advantage of preserving the semantic meaning of the sentence. They allowed swapping characters to create substitute words but if the word is out of the vocabulary then they repeated the last character to generate the substitute word.

\subsection{Multi-level Attack}

\begin{table}[ht]
\begin{tabular}{|c|c|c|}
\hline
Reference & Attack-type & Application    \\ \hline
Ebrahimi\shortcite{ebrahimi2017hotflip}  & White-box   & Classification \\ \hline
Blohm\shortcite{blohm2018comparing}   & \begin{tabular}[c]{@{}c@{}}White-box\\ and\\ Black-box\end{tabular} & \begin{tabular}[c]{@{}c@{}}Question-\\ Answering\end{tabular}                   \\ \hline
Wallace\shortcite{wallace2019universal} & White-box                                                           & \begin{tabular}[c]{@{}c@{}}Classification,\\ Question-\\ Answering\end{tabular} \\ \hline
\end{tabular}
\caption{Multi-level attack types with applications.}
\label{tab:my-table5}
\end{table}

HotFlip is a very popular, fast, and simple attack technique that was proposed by Ebrahimi \shortcite{ebrahimi2017hotflip}. This is a white-box gradient-based attack. In the core of the attack lies a simple flip operation which is based on the directional derivatives of the model with respect to one-hot encoding input. Only one forward and backward pass is required to predict the best flip operation. This attack can also include insertion and deletion as well if they are represented as character sequences. After estimating which changes ensure the highest classification errors a beam search algorithm finds a set of manipulations that works together to ensure the classifier is confused. Their original manuscript was on character level adversarial attack but they also showed that their approach can be extended to word-level attack as well. Since flipping a word to another has the possibility of losing its original value they flipped a word only if it satisfied certain conditions. They flipped a word if the cosine similarity of the word embeddings were higher than a given threshold and if they were members of the same parts of speech. They didn’t allow stop words to be removed.  

On the topic of question answering system Blohm \shortcite{blohm2018comparing} implemented word and sentence level white-box and black-box attacks. They started by achieving the state of the art score on the MovieQA dataset and then investigate different attacks effect. 

\begin{itemize}
    \item Word-level Black-box Attack: For this type of attack the authors substituted the words manually by choosing lexical substitutions that preserved their meanings. To ensure the words were inside of the vocabulary they only switched words which were included in the pretrained GloVe embeddings.
    \item Word-level White-box Attack: With the help of the attention model they used for classification they determined which sentence and which word was the most important one. This had a huge impact on the prediction results.
    \item Sentence-level Black-box Attack: Adopting the strategy of ADDANY attack proposed by \cite{jia2017adversarial} they initialized a sentence with ten common English words. Then each word is changed to another word which reduces the prediction confidence the most.
    \item Sentence-level White-box Attack: Similar to the word-level white-box attack they target the sentence which has the highest attention i.e. the plot sentence. They removed the plot sentence to see if the classifier was indeed focusing on it and its prediction capability.
\end{itemize}

Wallace \shortcite{wallace2019universal} proposed a technique in which they added tokens at the beginning or end of a sentence. They attempt to find universal adversarial triggers that are optimized based on the white-box approach but which can also be transferred to other models. At the very beginning, they start by choosing trigger lengths as this is an important criterion. Longer triggers are more effective but more noticeable than shorter ones. To replace the current tokens they took inspiration from the HotFlip approach proposed by \cite{ebrahimi2017hotflip}. They showed that for text classification tasks the triggers caused targeted errors for sentiment analysis and reading comprehension task triggers can cause paragraphs to generate arbitrary target prediction.

\section{Adversarial Defense}

As mentioned earlier most of the researchers focused on attacking DNN models in the field of NLP few focused on defending it. Here we divide our studied manuscripts into two sections one being the most common approach called adversarial training found in \cite{goodfellow2014explaining}. In the second section, we include all the research papers that try to tackle attacks by working on a specific defense technique.

\subsection{Adversarial Training}
Adversarial training is the process of training a model on both the original dataset and also adversarial example with correct labels. The general idea behind this approach is that, since the classifier model is now introduced to both original and adversarial data the model will now look beyond the perturbations and recognize the data properly.

Belinkov \shortcite{belinkov2017synthetic} in their experiments showed that training the model with different types of mixed noises improves the model's robustness to different kinds of noises. In the experiments of Li \shortcite{li2018textbugger} they also showed for TEXTBUGGER attack adversarial training can improve model performance and robustness against adversarial examples. In the experiments of Zang \shortcite{zang2019textual} they showed that their sememe based substitution and PSO based optimization improved classifiers' robustness to attacks. By using CharSwap during adversarial training on their attack Micheal showed that adversarial training can also improve the robustness of the model. Ebrahimi \shortcite{ebrahimi2017hotflip} in their manuscript of HotFlip also performed adversarial training. During their testing phase, they implemented beam search which wasn’t used for training hence the adversary in the training wasn’t strong as the testing ones. This reflects in their adversarial training experimental results as well. Though after training with adversarial examples the model attains certain robustness its accuracy isn’t as high as the original testing.

\subsection{Topic Specific Defense Techniques}
One of the major problems with adversarial training is that during training different types of attacks need to be known. Since adversaries don’t publicize their attack strategies adversarial training is limited by the users’ knowledge. If a user tries to perform adversarial training against all attacks known to him then the model would not be able to perform classification properly as it would have very low information on the original data.

In the research work of Alzanot \shortcite{alzantot2018generating} they found that their attack approach which was based on genetic algorithm was indifferent to adversarial training. A good reason for this would be that since their attack diversified the input so much adversarial training did not affect them. 

To protect models from synonym based attack techniques Wang \shortcite{wang2019natural} proposed the synonym encoding method (SEM) which puts an encoder network before the classifier model and checks for perturbations. In this approach, they cluster and encode all the synonyms to a unique code so that they force all the neighboring words to have similar codes in the embedding space. They compared their approach with adversarial training on four different attack techniques and showed that the SEM-based technique was better in the synonym substitution attack method.

Adversarial spelling mistakes were the prime concern of Pruthi \shortcite{pruthi2019combating} in their research work. Through their approach, they were able to handle adversarial examples which included insertion, deletion, swapping of characters, and keyboard mistakes. They used a semi character-based RNN model with three different back-off strategies for a word recognition model. They proposed three back-off strategies pass-through, back-off to a neutral word, back-off to the background model. They tested their approach against adversarial training and data augmentation based defense and found out that ScRNN with pass-through back-off strategy provided the highest robustness.

A defense framework was proposed by Zhou \shortcite{zhou2019learning} to determine whether a particular token is a perturbation or not. The discriminator provides some candidate perturbations and based on the candidate perturbations they used an embedding estimator to restore the original word and based on the context using the help of KNN search. The authors named this framework as DISP. This discriminator is trained on the original corpus during training time for figuring out which one is the perturbation. Token of the embedding corpus is fed to the embedding estimator to train it and recover the original word. During the testing phase, the discriminator provides candidate tokens which are perturbations, and for each of the candidate perturbation, the estimator provides an approximate embedding vector and attempts to restore the word. After this, the overall restored text can be passed to the model for prediction. To evaluate their frameworks' ability to identify perturbation tokens they compared their results against spell checking technique on three character level attacks and two-word level attacks. Results show that their approach was more successful in achieving better results. To test the robustness of their approach they compared against adversarial training, spell checking, and data augmentation. Their approach was able to perform in this experiment as well.

\section{Discussion}
We will be providing a discussion on some of the interesting findings that we found while studying different manuscripts. We are also going to shed some light on the challenges in this area. 

\subsection{Interesting Findings}

\begin{table*}[ht]
\begin{tabular}{|c|c|c|}
\hline
Task                                                                  & Dataset        & Reference                                                      \\ \hline
Classification &
  \begin{tabular}[c]{@{}c@{}}IMBD, SST-2, MR, AG News, \\ Yelp Review, DB-pedia, \\ Amazon Review, Trec07p, \\ Enron Spam Detection\end{tabular} &
  \begin{tabular}[c]{@{}c@{}}Gao\shortcite{gao2018black}, Li\shortcite{li2018textbugger}, Ebrahimi\shortcite{ebrahimi2017hotflip},\\  Kulesov\shortcite{kuleshovadversarial},Samanta\shortcite{samanta2017towards}, Zang\shortcite{zang2019textual}, \\   Liang\shortcite{liang2017deep}\end{tabular} \\ \hline
Machine Translation &
  \begin{tabular}[c]{@{}c@{}}Ted Talks parallel \\ corpus for IWSLT2016, \\ LDC corpus, NIST, \\ WMT'14\end{tabular} &
  \begin{tabular}[c]{@{}c@{}}Ebrahimi\shortcite{ebrahimi2018adversarial}, Belinkov\shortcite{belinkov2017synthetic}, \\ Cheng\shortcite{cheng2019robust}, Michel\shortcite{michel2019evaluation}\end{tabular} \\ \hline
\begin{tabular}[c]{@{}c@{}}Natural Language \\ Inference\end{tabular} & SNLI           & \begin{tabular}[c]{@{}c@{}}Alzantot\shortcite{alzantot2018generating}, Zang\shortcite{zang2019textual}, Zhao\shortcite{zhao2017generating}\end{tabular} \\ \hline
Question-Answering                                                    & SQuAD, MovieQA & Jia\shortcite{jia2017adversarial}, Blohm\shortcite{blohm2018comparing}                                                     \\ \hline
\end{tabular}
\caption{Dataset used by different researchers for attack generation }
\label{tab:my-table6}
\end{table*}

Based on the papers that we had studied we summarize and list out here some interesting findings.

\begin{itemize}
     \item Character-level Perturbations: It can be astounding to see that changing a single character can affect the model's prediction. Hosseini showed that adding dots(.) or space in words can be enough to confuse perspective api\cite{hosseini2017deceiving}. Not only this in HotFlip we have seen that the authors swapped a character based on the gradients to fool the model. So, while designing defense strategies a mere character level manipulation needs to be considered as well.
    \item Research Direction: From the papers that we studied we found out that most of the papers were based on different attack strategies. Very few papers were focused on defending the model. The same can be said for multi-stage attacks as well. Only a few researchers produced manuscripts for multi-stage attacks.
    \item Adversarial Example Generation: Through the studies of different manuscripts we found that to generate adversarial examples most of the researchers followed a two-step approach. The first being finding out which word was the most significant in providing prediction and the second step was to replace it with suitable candidates that benefit the adversary.

\end{itemize}

\subsection{Challenges}
In our study, we found several challenges in this field. They are mentioned below.

\begin{itemize}
    \item This phenomenon was first found in existence in the image domain and it has gained a lot of attention. Many research works have been published on it. It can be an easy assumption that we can use it in the text domain as well. However, there is a significant difference between them. In the image domain, the data is continuous but text data is discrete. Hence, the attacks proposed in the image domain cannot be utilized in text-domain.
    \item Another limitation of textual data is the concept of imperceptibility. In the image domain, the perturbation can often be made virtually imperceptible to human perception, causing humans and state-of-the-art models to disagree. However, in the text domain, small perturbations are usually clearly perceptible, and the replacement of a single word may drastically alter the semantics of the sentence and be noticeable to human beings. So, the structure of imperceptibility is an open issue.
    \item Till now no defense strategy can handle all different types of attacks that were mentioned here. Each defense strategy worked on a single type of attack approach. For example, for spelling mistakes, we can use the defense technique proposed by \cite{pruthi2019combating}. For synonym based attacks we can use the SEM model. A unified model that can tackle all these issues has not been proposed yet.
    \item The concept of universal perturbation has still not been introduced for textual data. In the image domain, researchers established a method that was able to generate a single perturbation that can fool the model. 
    \item Whenever a new attack technique is proposed researchers use different classifiers and datasets as there is no benchmark. From table 6 we can see for a particular application different datasets are used no ideal dataset is being used for attack generation. Since there is no benchmark it is not easy to compare different attack and defense strategies with each other. Lacking such a benchmark is a big gap in this field of research.
    \item There is no standard toolbox that can be used to easily reproduce different researchers' work. There are many toolbox which can be used in the image domain like cleverhans \cite{papernot2018cleverhans}, art \cite{art2018}, foolbox \cite{rauber2017foolbox} etc but there is no standard toolbox for text-domain.

\end{itemize}

\section{Conclusion}

In this review, we discussed the adversarial attack and defense techniques for textual data. Since the inception of adversarial examples, it has been a very important research topic for many aspects of deep learning applications. DNNs perform very well on a standard dataset but perform poorly in the presence of adversarial examples. We tried to present an accumulated view of why they exist, different attack and defense strategies based on their taxonomy. Also, we pointed out several challenges that can be tended to for getting future direction about research works in the future. 

\bibliography{acl2019}
\bibliographystyle{acl_natbib}

\end{document}